\documentclass[sigconf]{acmart}

\usepackage{graphicx}
\usepackage{fancyvrb}
\usepackage{color}
\usepackage[capitalize,noabbrev]{cleveref}
\usepackage{caption}
\usepackage{subcaption}

\AtBeginDocument{%
  \providecommand\BibTeX{{%
    \normalfont B\kern-0.5em{\scshape i\kern-0.25em b}\kern-0.8em\TeX}}}

\copyrightyear{2019}
\acmYear{2019}
\acmConference[CIKM '19]{The 28th ACM International Conference on Information and Knowledge Management}{November 3--7, 2019}{Beijing, China}
\acmBooktitle{The 28th ACM International Conference on Information and Knowledge Management (CIKM '19), November 3--7, 2019, Beijing, China}
\acmPrice{15.00}
\acmDOI{10.1145/3357384.3357860}
\acmISBN{978-1-4503-6976-3/19/11}

\makeatletter
\def\PY@reset{\let\PY@it=\relax \let\PY@bf=\relax%
    \let\PY@ul=\relax \let\PY@tc=\relax%
    \let\PY@bc=\relax \let\PY@ff=\relax}
\def\PY@tok#1{\csname PY@tok@#1\endcsname}
\def\PY@toks#1+{\ifx\relax#1\empty\else%
    \PY@tok{#1}\expandafter\PY@toks\fi}
\def\PY@do#1{\PY@bc{\PY@tc{\PY@ul{%
    \PY@it{\PY@bf{\PY@ff{#1}}}}}}}
\def\PY#1#2{\PY@reset\PY@toks#1+\relax+\PY@do{#2}}

\expandafter\def\csname PY@tok@w\endcsname{\def\PY@tc##1{\textcolor[rgb]{0.73,0.73,0.73}{##1}}}
\expandafter\def\csname PY@tok@c\endcsname{\let\PY@it=\textit\def\PY@tc##1{\textcolor[rgb]{0.25,0.50,0.50}{##1}}}
\expandafter\def\csname PY@tok@cp\endcsname{\def\PY@tc##1{\textcolor[rgb]{0.74,0.48,0.00}{##1}}}
\expandafter\def\csname PY@tok@k\endcsname{\let\PY@bf=\textbf\def\PY@tc##1{\textcolor[rgb]{0.00,0.50,0.00}{##1}}}
\expandafter\def\csname PY@tok@kp\endcsname{\def\PY@tc##1{\textcolor[rgb]{0.00,0.50,0.00}{##1}}}
\expandafter\def\csname PY@tok@kt\endcsname{\def\PY@tc##1{\textcolor[rgb]{0.69,0.00,0.25}{##1}}}
\expandafter\def\csname PY@tok@o\endcsname{\def\PY@tc##1{\textcolor[rgb]{0.40,0.40,0.40}{##1}}}
\expandafter\def\csname PY@tok@ow\endcsname{\let\PY@bf=\textbf\def\PY@tc##1{\textcolor[rgb]{0.67,0.13,1.00}{##1}}}
\expandafter\def\csname PY@tok@nb\endcsname{\def\PY@tc##1{\textcolor[rgb]{0.00,0.50,0.00}{##1}}}
\expandafter\def\csname PY@tok@nf\endcsname{\def\PY@tc##1{\textcolor[rgb]{0.00,0.00,1.00}{##1}}}
\expandafter\def\csname PY@tok@nc\endcsname{\let\PY@bf=\textbf\def\PY@tc##1{\textcolor[rgb]{0.00,0.00,1.00}{##1}}}
\expandafter\def\csname PY@tok@nn\endcsname{\let\PY@bf=\textbf\def\PY@tc##1{\textcolor[rgb]{0.00,0.00,1.00}{##1}}}
\expandafter\def\csname PY@tok@ne\endcsname{\let\PY@bf=\textbf\def\PY@tc##1{\textcolor[rgb]{0.82,0.25,0.23}{##1}}}
\expandafter\def\csname PY@tok@nv\endcsname{\def\PY@tc##1{\textcolor[rgb]{0.10,0.09,0.49}{##1}}}
\expandafter\def\csname PY@tok@no\endcsname{\def\PY@tc##1{\textcolor[rgb]{0.53,0.00,0.00}{##1}}}
\expandafter\def\csname PY@tok@nl\endcsname{\def\PY@tc##1{\textcolor[rgb]{0.63,0.63,0.00}{##1}}}
\expandafter\def\csname PY@tok@ni\endcsname{\let\PY@bf=\textbf\def\PY@tc##1{\textcolor[rgb]{0.60,0.60,0.60}{##1}}}
\expandafter\def\csname PY@tok@na\endcsname{\def\PY@tc##1{\textcolor[rgb]{0.49,0.56,0.16}{##1}}}
\expandafter\def\csname PY@tok@nt\endcsname{\let\PY@bf=\textbf\def\PY@tc##1{\textcolor[rgb]{0.00,0.50,0.00}{##1}}}
\expandafter\def\csname PY@tok@nd\endcsname{\def\PY@tc##1{\textcolor[rgb]{0.67,0.13,1.00}{##1}}}
\expandafter\def\csname PY@tok@s\endcsname{\def\PY@tc##1{\textcolor[rgb]{0.73,0.13,0.13}{##1}}}
\expandafter\def\csname PY@tok@sd\endcsname{\let\PY@it=\textit\def\PY@tc##1{\textcolor[rgb]{0.73,0.13,0.13}{##1}}}
\expandafter\def\csname PY@tok@si\endcsname{\let\PY@bf=\textbf\def\PY@tc##1{\textcolor[rgb]{0.73,0.40,0.53}{##1}}}
\expandafter\def\csname PY@tok@se\endcsname{\let\PY@bf=\textbf\def\PY@tc##1{\textcolor[rgb]{0.73,0.40,0.13}{##1}}}
\expandafter\def\csname PY@tok@sr\endcsname{\def\PY@tc##1{\textcolor[rgb]{0.73,0.40,0.53}{##1}}}
\expandafter\def\csname PY@tok@ss\endcsname{\def\PY@tc##1{\textcolor[rgb]{0.10,0.09,0.49}{##1}}}
\expandafter\def\csname PY@tok@sx\endcsname{\def\PY@tc##1{\textcolor[rgb]{0.00,0.50,0.00}{##1}}}
\expandafter\def\csname PY@tok@m\endcsname{\def\PY@tc##1{\textcolor[rgb]{0.40,0.40,0.40}{##1}}}
\expandafter\def\csname PY@tok@gh\endcsname{\let\PY@bf=\textbf\def\PY@tc##1{\textcolor[rgb]{0.00,0.00,0.50}{##1}}}
\expandafter\def\csname PY@tok@gu\endcsname{\let\PY@bf=\textbf\def\PY@tc##1{\textcolor[rgb]{0.50,0.00,0.50}{##1}}}
\expandafter\def\csname PY@tok@gd\endcsname{\def\PY@tc##1{\textcolor[rgb]{0.63,0.00,0.00}{##1}}}
\expandafter\def\csname PY@tok@gi\endcsname{\def\PY@tc##1{\textcolor[rgb]{0.00,0.63,0.00}{##1}}}
\expandafter\def\csname PY@tok@gr\endcsname{\def\PY@tc##1{\textcolor[rgb]{1.00,0.00,0.00}{##1}}}
\expandafter\def\csname PY@tok@ge\endcsname{\let\PY@it=\textit}
\expandafter\def\csname PY@tok@gs\endcsname{\let\PY@bf=\textbf}
\expandafter\def\csname PY@tok@gp\endcsname{\let\PY@bf=\textbf\def\PY@tc##1{\textcolor[rgb]{0.00,0.00,0.50}{##1}}}
\expandafter\def\csname PY@tok@go\endcsname{\def\PY@tc##1{\textcolor[rgb]{0.53,0.53,0.53}{##1}}}
\expandafter\def\csname PY@tok@gt\endcsname{\def\PY@tc##1{\textcolor[rgb]{0.00,0.27,0.87}{##1}}}
\expandafter\def\csname PY@tok@err\endcsname{\def\PY@bc##1{\setlength{\fboxsep}{0pt}\fcolorbox[rgb]{1.00,0.00,0.00}{1,1,1}{\strut ##1}}}
\expandafter\def\csname PY@tok@kc\endcsname{\let\PY@bf=\textbf\def\PY@tc##1{\textcolor[rgb]{0.00,0.50,0.00}{##1}}}
\expandafter\def\csname PY@tok@kd\endcsname{\let\PY@bf=\textbf\def\PY@tc##1{\textcolor[rgb]{0.00,0.50,0.00}{##1}}}
\expandafter\def\csname PY@tok@kn\endcsname{\let\PY@bf=\textbf\def\PY@tc##1{\textcolor[rgb]{0.00,0.50,0.00}{##1}}}
\expandafter\def\csname PY@tok@kr\endcsname{\let\PY@bf=\textbf\def\PY@tc##1{\textcolor[rgb]{0.00,0.50,0.00}{##1}}}
\expandafter\def\csname PY@tok@bp\endcsname{\def\PY@tc##1{\textcolor[rgb]{0.00,0.50,0.00}{##1}}}
\expandafter\def\csname PY@tok@fm\endcsname{\def\PY@tc##1{\textcolor[rgb]{0.00,0.00,1.00}{##1}}}
\expandafter\def\csname PY@tok@vc\endcsname{\def\PY@tc##1{\textcolor[rgb]{0.10,0.09,0.49}{##1}}}
\expandafter\def\csname PY@tok@vg\endcsname{\def\PY@tc##1{\textcolor[rgb]{0.10,0.09,0.49}{##1}}}
\expandafter\def\csname PY@tok@vi\endcsname{\def\PY@tc##1{\textcolor[rgb]{0.10,0.09,0.49}{##1}}}
\expandafter\def\csname PY@tok@vm\endcsname{\def\PY@tc##1{\textcolor[rgb]{0.10,0.09,0.49}{##1}}}
\expandafter\def\csname PY@tok@sa\endcsname{\def\PY@tc##1{\textcolor[rgb]{0.73,0.13,0.13}{##1}}}
\expandafter\def\csname PY@tok@sb\endcsname{\def\PY@tc##1{\textcolor[rgb]{0.73,0.13,0.13}{##1}}}
\expandafter\def\csname PY@tok@sc\endcsname{\def\PY@tc##1{\textcolor[rgb]{0.73,0.13,0.13}{##1}}}
\expandafter\def\csname PY@tok@dl\endcsname{\def\PY@tc##1{\textcolor[rgb]{0.73,0.13,0.13}{##1}}}
\expandafter\def\csname PY@tok@s2\endcsname{\def\PY@tc##1{\textcolor[rgb]{0.73,0.13,0.13}{##1}}}
\expandafter\def\csname PY@tok@sh\endcsname{\def\PY@tc##1{\textcolor[rgb]{0.73,0.13,0.13}{##1}}}
\expandafter\def\csname PY@tok@s1\endcsname{\def\PY@tc##1{\textcolor[rgb]{0.73,0.13,0.13}{##1}}}
\expandafter\def\csname PY@tok@mb\endcsname{\def\PY@tc##1{\textcolor[rgb]{0.40,0.40,0.40}{##1}}}
\expandafter\def\csname PY@tok@mf\endcsname{\def\PY@tc##1{\textcolor[rgb]{0.40,0.40,0.40}{##1}}}
\expandafter\def\csname PY@tok@mh\endcsname{\def\PY@tc##1{\textcolor[rgb]{0.40,0.40,0.40}{##1}}}
\expandafter\def\csname PY@tok@mi\endcsname{\def\PY@tc##1{\textcolor[rgb]{0.40,0.40,0.40}{##1}}}
\expandafter\def\csname PY@tok@il\endcsname{\def\PY@tc##1{\textcolor[rgb]{0.40,0.40,0.40}{##1}}}
\expandafter\def\csname PY@tok@mo\endcsname{\def\PY@tc##1{\textcolor[rgb]{0.40,0.40,0.40}{##1}}}
\expandafter\def\csname PY@tok@ch\endcsname{\let\PY@it=\textit\def\PY@tc##1{\textcolor[rgb]{0.25,0.50,0.50}{##1}}}
\expandafter\def\csname PY@tok@cm\endcsname{\let\PY@it=\textit\def\PY@tc##1{\textcolor[rgb]{0.25,0.50,0.50}{##1}}}
\expandafter\def\csname PY@tok@cpf\endcsname{\let\PY@it=\textit\def\PY@tc##1{\textcolor[rgb]{0.25,0.50,0.50}{##1}}}
\expandafter\def\csname PY@tok@c1\endcsname{\let\PY@it=\textit\def\PY@tc##1{\textcolor[rgb]{0.25,0.50,0.50}{##1}}}
\expandafter\def\csname PY@tok@cs\endcsname{\let\PY@it=\textit\def\PY@tc##1{\textcolor[rgb]{0.25,0.50,0.50}{##1}}}

\def\PYZob{\char`\{}
\def\PYZcb{\char`\}}

\def\PYZdq{\char`\"}

\makeatother

\begin{document}
\fancyhead{}
\title{Model Asset eXchange: Path to Ubiquitous Deep Learning Deployment}

\newcommand{\ouraffiliation}{
\affiliation{%
  \institution{Center for Open-Source Data \& AI Technologies (CODAIT), IBM}
  \streetaddress{505 Howard St}
  \city{San Francisco}
  \state{California}
  \country{USA}
  \postcode{94105}
}
}

\author{Alex Bozarth}
\ouraffiliation{}
\email{ajbozart@us.ibm.com}
\author{Brendan Dwyer}
\ouraffiliation{}
\email{Brendan.Dwyer@ibm.com}
\author{Fei Hu}
\ouraffiliation{}
\email{Fei.Hu1@ibm.com}
\author{Daniel Jalova}
\ouraffiliation{}
\email{djalova@us.ibm.com}
\author{Karthik Muthuraman}
\ouraffiliation{}
\email{Karthik.Muthuraman@ibm.com}
\author{Nick Pentreath}
\ouraffiliation{}
\email{NickP@za.ibm.com}
\author{Simon Plovyt}
\ouraffiliation{}
\email{simon@ibm.com}
\author{Gabriela de Queiroz}
\ouraffiliation{}
\email{gdq@ibm.com}
\author{Saishruthi Swaminathan}
\ouraffiliation{}
\email{saishruthi.tn@ibm.com}
\author{Patrick Titzler}
\ouraffiliation{}
\email{ptitzler@us.ibm.com}
\author{Xin Wu}
\ouraffiliation{}
\email{xinwu@us.ibm.com}
\author{Hong Xu}
\authornote{All authors contributed equally. Please contact this author or visit \url{http://ibm.biz/max-slack} for questions regarding this paper.}
\orcid{0000-0001-7874-4518}
\ouraffiliation{}
\email{hongx@ibm.com}
\author{Frederick R Reiss}
\ouraffiliation{}
\additionalaffiliation{%
  \institution{IBM Research}
  \city{San Jose}
  \state{California}
  \country{USA}
}
\email{frreiss@us.ibm.com}
\author{Vijay Bommireddipalli}
\ouraffiliation{}
\email{vijayrb@us.ibm.com}

\renewcommand{\shortauthors}{Bozarth et al.}

\begin{abstract}
  A recent trend observed in traditionally challenging fields such as
  computer vision and natural language processing has been the
  significant performance gains shown by deep learning (DL). In many
  different research fields, DL models have been evolving rapidly and
  become ubiquitous. Despite researchers' excitement, unfortunately,
  most software developers are not DL experts and oftentimes have a
  difficult time following the booming DL research outputs. As a result,
  it usually takes a significant amount of time for the latest superior
  DL models to prevail in industry. This issue is further exacerbated by
  the common use of sundry incompatible DL programming frameworks, such
  as Tensorflow, PyTorch, Theano, etc. To address this issue, we propose
  a system, called Model Asset Exchange (MAX), that avails developers of
  easy access to state-of-the-art DL models. Regardless of the
  underlying DL programming frameworks, it provides an open source
  Python library (called the MAX framework) that wraps DL models and
  unifies programming interfaces with our standardized RESTful APIs.
  These RESTful APIs enable developers to exploit the wrapped DL models
  for inference tasks without the need to fully understand different DL
  programming frameworks. Using MAX, we have wrapped and open-sourced
  more than 30 state-of-the-art DL models from various research fields,
  including computer vision, natural language processing and signal
  processing, etc. In the end, we selectively demonstrate two web
  applications that are built on top of MAX, as well as the process of
  adding a DL model to MAX\@.
\end{abstract}


\maketitle

\section{Introduction}

\begin{figure*}[t]
    \includegraphics[width=\linewidth]{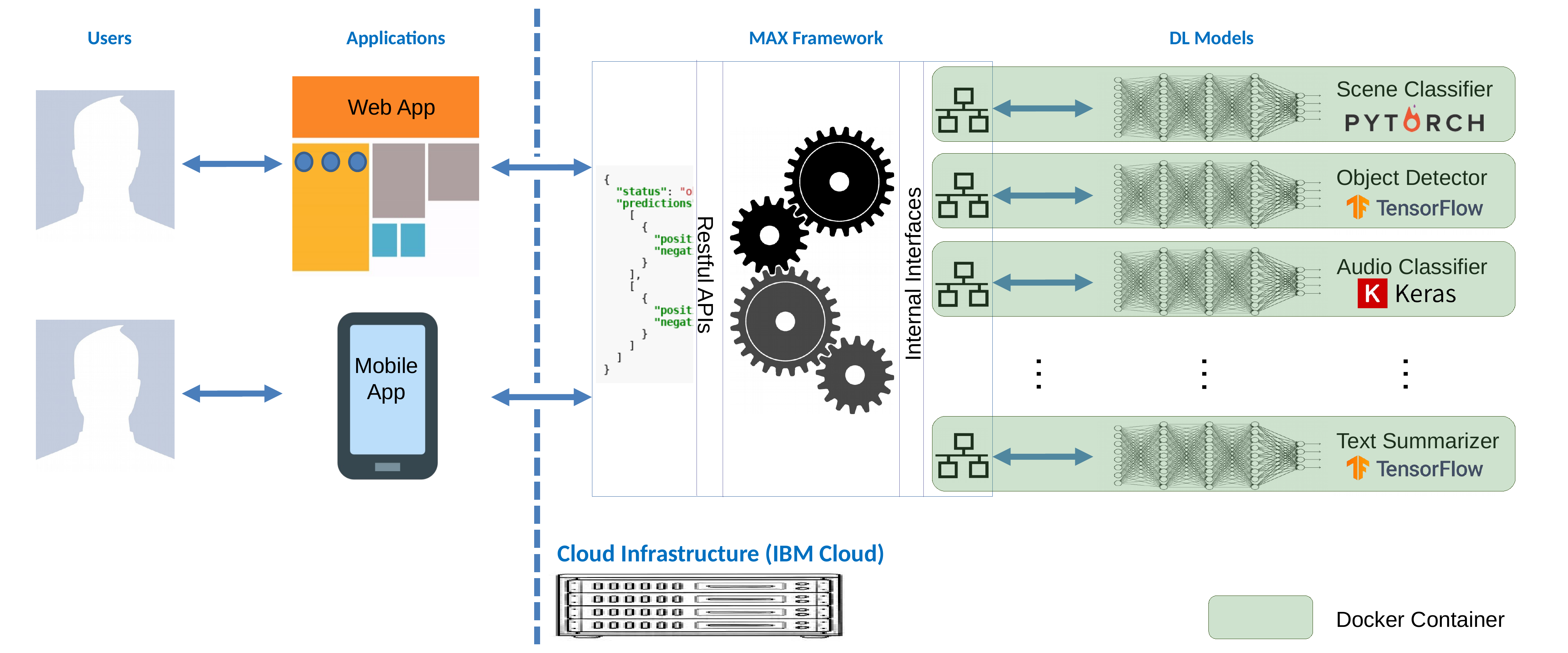}
    \caption{Design of the MAX architecture. The software components to
      the right of the \textit{vertical blue dashed line} are on IBM
      Cloud. Software components inside \textit{the round-cornered green
        rectangles} run in Docker containers.}
    \vspace{-1.5em}
    \caption*{\tiny \normalfont Source of some subcomponents of the figure:
      \url{https://upload.wikimedia.org/wikipedia/commons/3/38/Ethernet\_Port.svg}
      , https://commons.wikimedia.org/wiki/File:Gears.png , https://commons.wikimedia.org/wiki/File:Icons8\_flat\_phone\_android.svg}
  \label{fig:max}
\end{figure*}

In the past decade, research fields of \textit{artificial intelligence}
(AI) have advanced drastically. Most noticeably, \textit{deep learning}
(DL) has gained dramatic performance improvements and revolutionized the
whole research field of AI. On the one hand, for many tasks that were
considered ``impossible'' for AI to achieve a performance comparable to
human, such as playing Go~\cite{silver2016} and image
recognition~\cite{he2015}, DL has demonstrated its power by surpassing human
performance with a large margin. On the other hand, DL has also become
ubiquitous in many areas, especially in Internet of things. Therefore,
it is essential to systematize DL models.

DL researchers have been celebrating their remarkable and laudable
achievements: Some DL models that arose from applied DL research fields
such as computer vision and natural language processing, have been
revolutionizing these fields. However, the industry has been struggling
to make use of them for the following reasons:

\textbf{Booming research results} Every year, thousands of new DL
research results are published. The state of the art is pushed forward
rapidly in many DL research fields. For a non-DL expert, it usually
takes a significant amount of time to implement necessary blocks derived
from research papers.

\textbf{Mathematical complexity} Sometimes DL models require decent
knowledge on statistics and linear algebra to grasp. This mathematical
barriers are often insurmountable for most software developers due to
the diversity in their backgrounds.

\textbf{Confusing terminologies} New terminologies have been emerging
due to DL's rapid evolution and wide adoption in a variety of research
fields. Unfortunately, these new terminologies are often inconsistent
and thus frequently confuse non-DL experts.

These issues have been severely impeding the adoption of DL in industry.
Therefore, to fill this gap, it is imperative to develop a software
system that reduces the required minimum knowledge to get started with
DL models.

In addition, the difficulty to harness DL in industry is further
exacerbated by the common use of sundry DL programming frameworks,
including Tensorflow~\cite{tensorflow}, PyTorch~\cite{pytorch},
Theano~\cite{theano}, etc. These DL programming frameworks are
incompatible with each other and their software structures are usually
fundamentally different. Hence, it usually requires strenuous effort to
port programming code based on one DL programming framework to another.
Therefore, a software system that unifies these programming
interfaces and blocks is critical to the success of DL in industry.

In this demo, we present a software system, called \textit{Model Asset
  eXchange}
(MAX)\footnote{\url{https://developer.ibm.com/exchanges/models/}}, that
addresses aforementioned difficulties. The paper is organized as
follows. In \cref{sec:architecture}, we present the architecture and
software components of MAX\@. In \cref{sec:demonstration}, we describe
our demonstration.

\section{Architecture Design and Software Components of MAX}
\label{sec:architecture}

In this section, we describe our architecture design and various
software components of MAX.

\subsection{Architecture Design}

MAX is a software system that employs an extensible and distributive
architecture and makes use of state-of-the-art container technology and
cloud infrastructures. \Cref{fig:max} illustrates its architecture. MAX
is hosted on a cloud infrastructure, such as IBM cloud, and communicates
with web applications via standardized RESTful APIs. It is undergirded
by a powerful abstract component, named the MAX framework. The MAX
framework wraps DL models implemented in different DL programming
frameworks and provides programming interfaces in an uniform style,
which effectively enables developers to use DL models without the need
to dive into DL programming frameworks. Each implementation of DL model
runs in isolated \textit{Docker containers}, which promotes security and
effectively turns the architecture to be easily distributive and
extensible. Additionally, we build MAX exclusively on top of open source
technologies, which promotes the open and collaborative culture that
academia generally embraces.

\subsection{Software Components}

In this subsection, we describe MAX's various software components as
shown in \cref{fig:max}.

\subsubsection{The MAX Framework}

The MAX framework is a Python library that wraps DL models to unify the
programming interface. To wrap a model, it simply requires implementing
functions that process input and output. This simplicity is key to the
MAX framework's agnosticism to DL programming frameworks.

\subsubsection{DL Models} MAX can accommodate DL models written in
different DL programming frameworks. The MAX framework communicates with
DL models via standardized Python programming interfaces. To use a DL
model in MAX, we only need to adapt its Python programming interface,
i.e., wrap the DL model. Once the DL model is wrapped, it is available
throughout the whole MAX system and does not require further adaptation
in the future. This Python programming interface is objected-oriented:
Wrapping only requires inheriting specific classes and implementing some
predefined class functions by converting input and output of DL
models to data structures acceptable to the MAX framework.

The wrapped DL models and their programming interface with the MAX
framework are hosted in \textit{Docker
  containers}\footnote{\url{https://www.docker.com}}. A container is an
isolated instance of environment that hosts software of interest and its
runtime. This isolation in general promotes extensibility,
distributability, and security.

For example, without the help of containers, it is usually difficult to
deploy two DL models depending on conflicting runtime environments (such
as different versions of TensorFlow) on the same hosting OS in the same
physical computer. Containers solve this issue by creating an isolated
virtual runtime environment for each DL model. For another example, if
multiple DL models are deployed on the same hosting OS directly, a
security vulnerability in one DL model would also normally risk other DL
models.

Additionally, MAX uses Docker containers instead of traditional
isolation/container technologies such as physical isolation (multiple
physical computers) or virtual machines (emulated hardware environments
running on host OSes). The reason is that traditional
isolation/container technologies are in general computationally costly,
as each isolated node runs a complete \textit{operating system} (OS)
instance, either physical or virtual.

To mitigate this issue, Docker containers only cost moderate
computational resources while retaining a high degree of isolation:
Based on Linux container technologies, Docker containers share one
single OS kernel instance that is the same as the one that steers the
host OS. Therefore, by employing Docker containers, MAX is able to
operate under low computational cost with very little compromise in
extensibility, distributability, or security that isolation/container
technologies provide.

\subsubsection{RESTful APIs: Between Applications and the MAX Framework}

MAX provides a standardized DL programming framework-agnostic
programming interface as RESTful APIs, which effectively avails
developers of DL models without requiring them to dive into the
various DL programming frameworks.

For each DL model, MAX's output is in the JSON format following a
standardized specification. This standardization enables developers to
quickly adapt their applications by replacing the underlying DL model
with very little and often zero modification to the code that interacts
with the DL model. This is in sharp contrast with the current common
practice: Due to the non-standardized programming interfaces, when
replacing underlying DL models, developers usually have to drastically
modify their code and frequently find themselves mired in figuring out
the correct usage of often abstrusely defined APIs. MAX also integrates
Swagger\footnote{\url{https://swagger.io/}} to make a \textit{graphical
  user interface} (GUI) automatically available to all wrapped DL
models. An example is shown
below\footnote{Taken from
  \url{https://github.com/IBM/MAX-Text-Sentiment-Classifier} .}:

  \begin{Verbatim}[fontsize=\scriptsize,commandchars=\\\{\}]
\PY{p}{\PYZob{}}
  \PY{n+nt}{\PYZdq{}status\PYZdq{}}\PY{p}{:} \PY{l+s+s2}{\PYZdq{}ok\PYZdq{}}\PY{p}{,}
  \PY{n+nt}{\PYZdq{}predictions\PYZdq{}}\PY{p}{:}\PY{p}{[}
    \PY{p}{[}\PY{p}{\PYZob{}}\PY{n+nt}{\PYZdq{}positive\PYZdq{}}\PY{p}{:} \PY{l+m+mf}{0.9977352619171143}\PY{p}{,} \PY{n+nt}{\PYZdq{}negative\PYZdq{}}\PY{p}{:} \PY{l+m+mf}{0.002264695707708597}\PY{p}{\PYZcb{}}\PY{p}{]}\PY{p}{,}
    \PY{p}{[}\PY{p}{\PYZob{}}\PY{n+nt}{\PYZdq{}positive\PYZdq{}}\PY{p}{:} \PY{l+m+mf}{0.001138084102421999}\PY{p}{,} \PY{n+nt}{\PYZdq{}negative\PYZdq{}}\PY{p}{:} \PY{l+m+mf}{0.9988619089126587}\PY{p}{\PYZcb{}}\PY{p}{]}
  \PY{p}{]}
\PY{p}{\PYZcb{}}
  \end{Verbatim}

\section{Demonstration}
\label{sec:demonstration}

\begin{figure}[t]
  \begin{subfigure}[t]{0.79\linewidth}
    \centering
    \includegraphics[width=\linewidth]{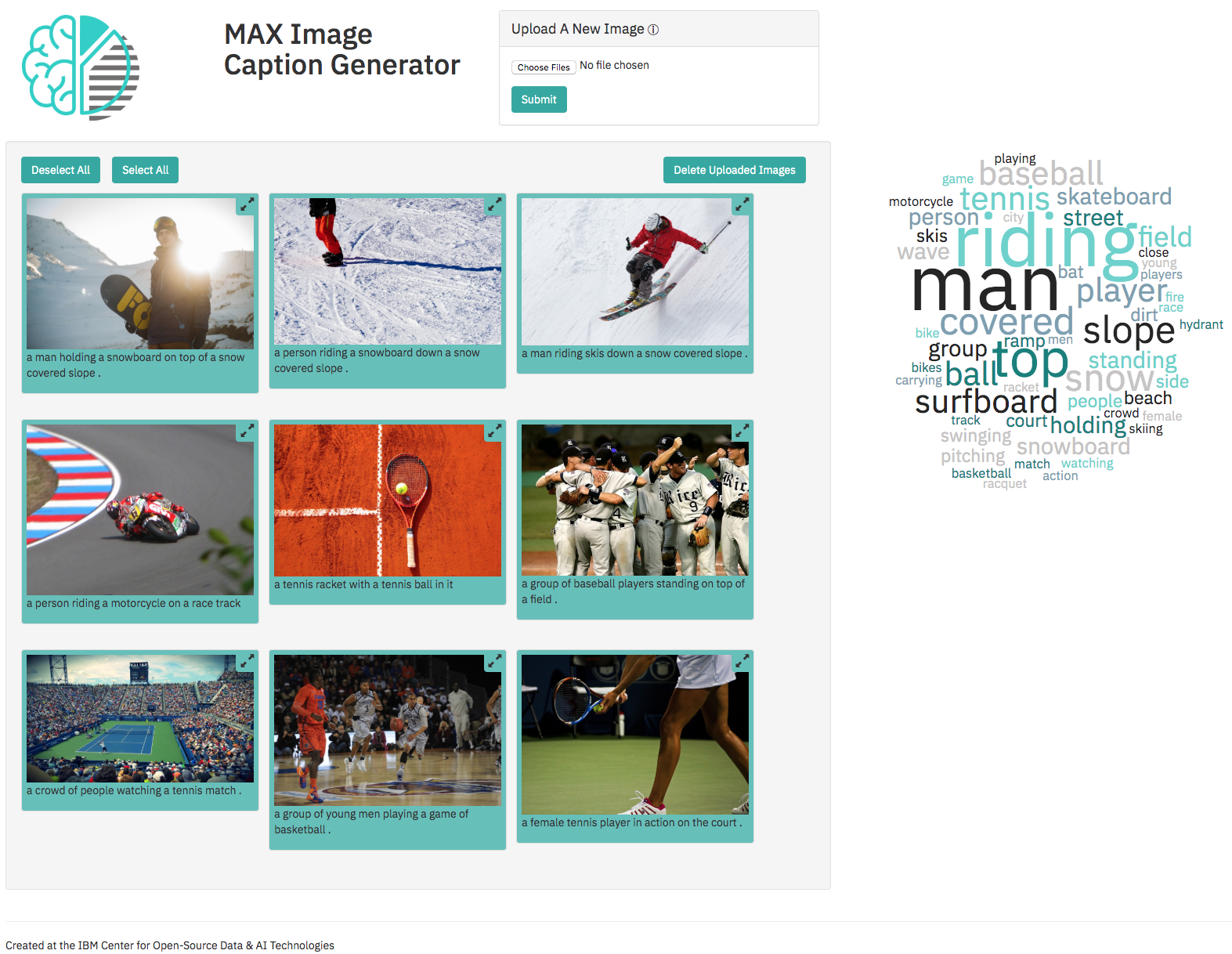}
    \caption{Image caption generator (web UI screenshot)}
    \label{fig:img_caption}
  \end{subfigure}
  \hfill
  \begin{subfigure}[t]{0.79\linewidth}
    \centering
    \includegraphics[width=\linewidth]{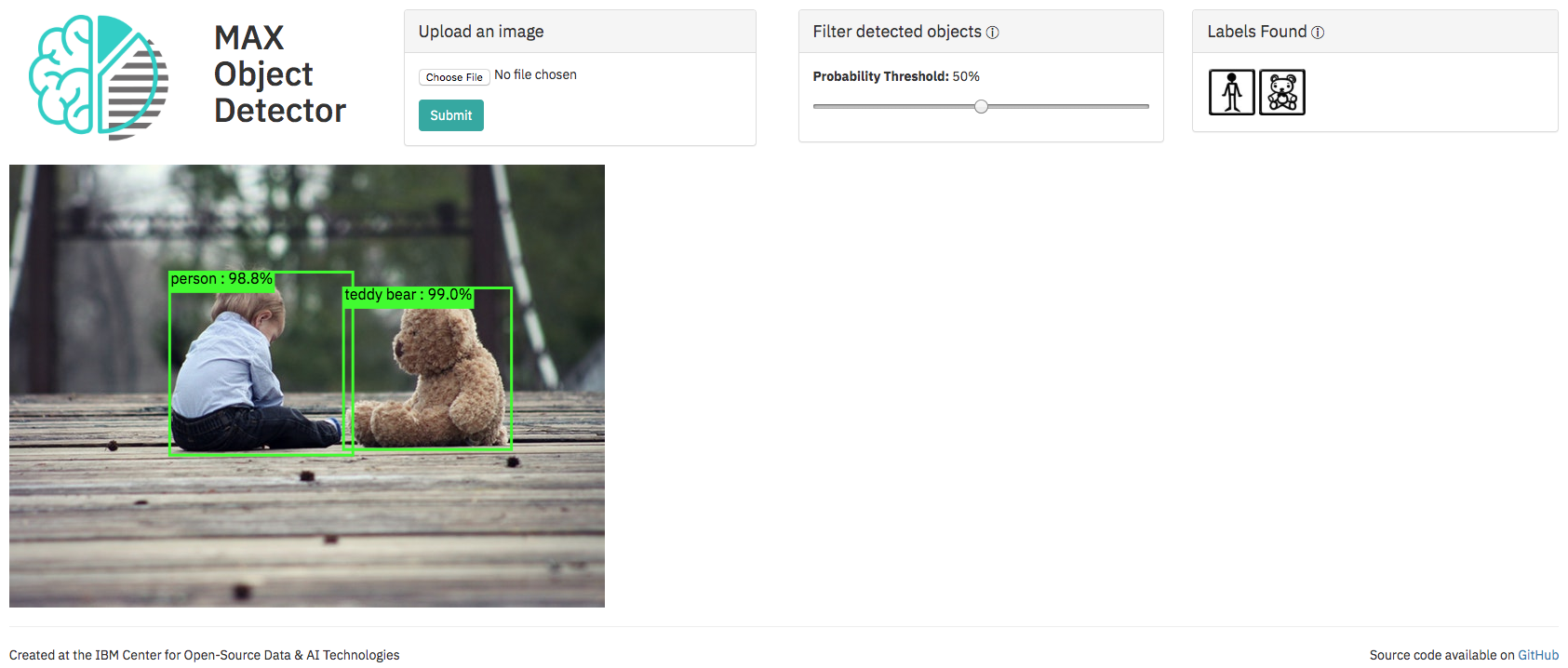}
    \caption{Object detector (web UI screenshot)}
    \label{fig:object_detector}
  \end{subfigure}
  \hfill
  \caption{Web application demonstration.}
  \label{fig:web_app}
\end{figure}

In this section, we describe our demonstration.

\subsection{Web Applications Built upon MAX}

In this subsection, we describe our demonstration of the architecture of
MAX using two web applications, an object
detector~\cite{huang2017,lin2014,liu2016,howard2017} and an image
caption generator~\cite{vinyals2017}, that are built on top of MAX\@. We
describe in turn the demonstration of software components (as shown in
\cref{fig:max}) and interfaces between them.

\begin{figure}[t]
  \centering
  \includegraphics[width=0.7\linewidth]{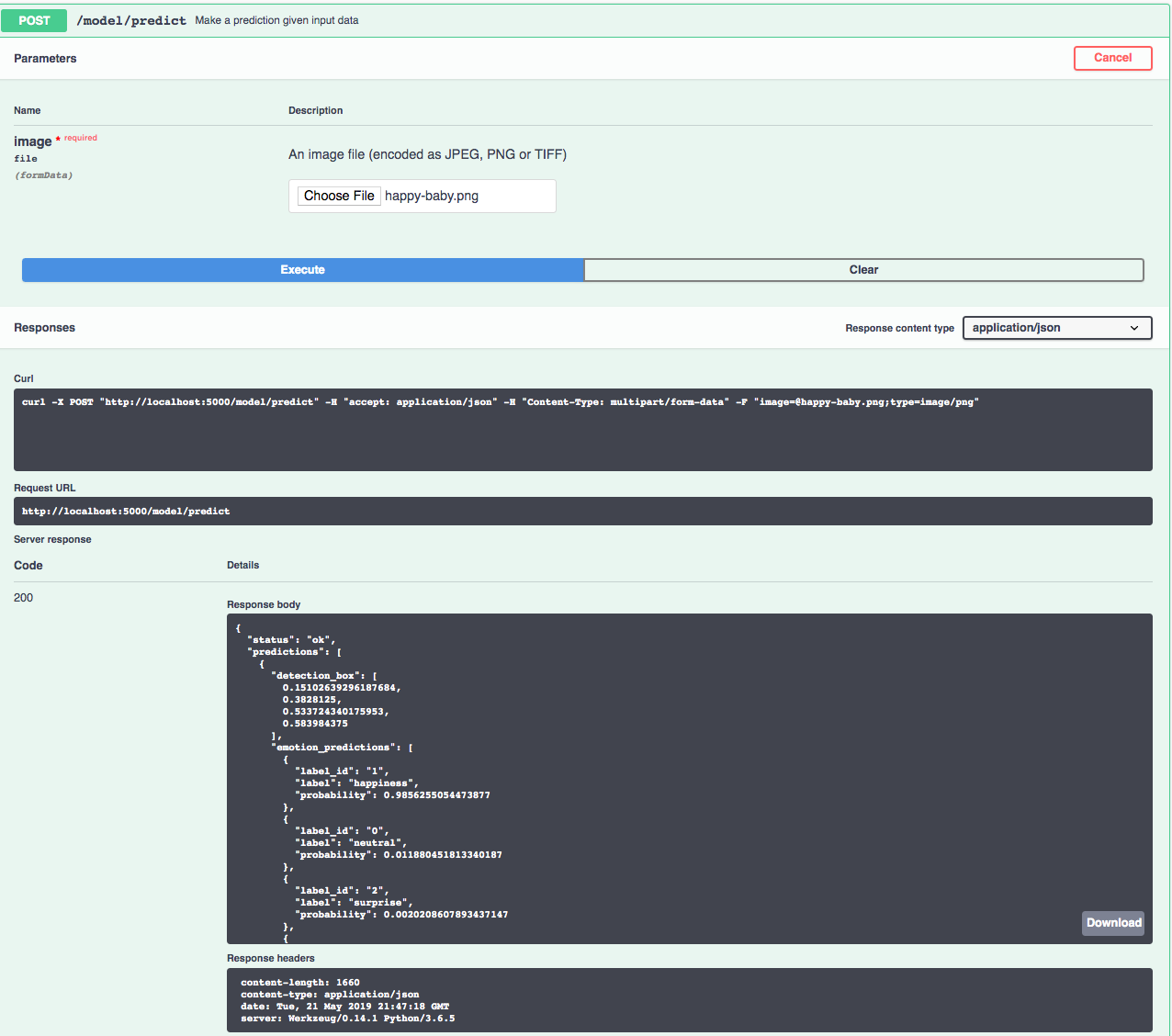}
  \caption{Example JSON output from the image caption generator.}
  \label{fig:json_demo}
\end{figure}

\begin{itemize}
\item \textbf{Web UI:} They interact with users via GUIs and communicate
  with the MAX framework via a RESTful JSON interface.
  \Cref{fig:object_detector,fig:img_caption} preview our demonstration.
\item \textbf{JSON Interface Between Web UI and the MAX Framework:} This
  interface is RESTful and based on the JSON format. Upon users'
  requests, the Web UI accordingly sends the MAX framework a
  JSON-formatted string following a predefined specification.
  \Cref{fig:json_demo} previews our demonstration (using Swagger) of
  this JSON interface.
\item \textbf{Python Programming Interface Between DL Models and the MAX
    Framework:} We will demonstrate the Python interface that bridges
  these two components.
\end{itemize}

\subsection{Adding a DL Model to MAX}

We will also interactively demonstrate the process to add a DL model to
MAX and thus make it available to MAX users. This includes three steps:
(1) Wrapping a DL model using the MAX framework, (2) building a Docker
image that hosts the wrapped DL model, and (3) optionally uploading to
IBM Cloud. Additionally, for this process, we have also created a
skeleton called
MAX-Skeleton\footnote{\url{https://github.com/IBM/MAX-Skeleton}} as a
convenient starting point for typical use cases.

\section{Conclusion}

In this demo paper, to address the difficulties in the industry's
adoption of DL models from DL research fields, we presented MAX, a
software system that employs an extensible and distributive architecture
and makes use of state-of-the-art container technology and cloud
infrastructures. In particular, we described the architecture and
software components of MAX as well as the interaction between them in
detail. Finally, we proposed our demonstration of two web applications
that are built on top of MAX, as well as the process of adding a DL
model to MAX\@.

\bibliographystyle{ACM-Reference-Format}
\bibliography{refs}
\end{document}